\documentclass[10pt,twocolumn,letterpaper]{article}

\usepackage{btas}
\usepackage{times}
\usepackage{epsfig}
\usepackage{graphicx}
\usepackage{amsmath}
\usepackage{amssymb}
\usepackage{subfigure}
\usepackage{fancyhdr}
\btasfinalcopy % *** Uncomment this line for the final submission

\ifbtasfinal\fi
\begin{document}

%%%%%%%%% TITLE
\title{Adversarial Occlusion-aware Face Detection}

\author{Yujia Chen$^{1,2}$, Lingxiao Song$^{1,2,3}$, Ran He$^{1,2,3}$\thanks{corresponding author}\\
$^1$National Laboratory of Pattern Recognition, CASIA\\
$^2$Center for Research on Intelligent Perception and Computing, CASIA\\
$^3$University of Chinese Academy of Sciences, Beijing 100190, China
}

%\maketitle
%\thispagestyle{empty}

\maketitle
\thispagestyle{fancy}
%\fancyhead{}
%\lhead{}
%\lfoot{978-1-5386-7180-1/18/\$31.00 $\copyright$2018 IEEE}
%\cfoot{}
%\rfoot{}

%%%%%%%%% ABSTRACT
\begin{abstract}
Occluded face detection is a challenging detection task due to the large appearance variations incurred by various real-world occlusions. This paper introduces an Adversarial Occlusion-aware Face Detector (AOFD) by simultaneously detecting occluded faces and segmenting occluded areas. Specifically, we employ an adversarial training strategy to generate occlusion-like face features that are difficult for a face detector to recognize. Occlusion is predicted simultaneously while detecting occluded faces and the occluded area is utilized as an auxiliary instead of being regarded as a hindrance. Moreover, the supervisory signals from the segmentation branch will reversely affect the features, helping extract more informative features. Consequently, AOFD is able to find the faces with few exposed facial landmarks with very high confidences and keeps high detection accuracy even for masked faces. Extensive experiments demonstrate that AOFD not only significantly outperforms state-of-the-art methods on the MAFA occluded face detection dataset, but also achieves competitive detection accuracy on benchmark dataset for general face detection such as FDDB.
\end{abstract}

%%%%%%%%% BODY TEXT

\section{Introduction}

Face detection has been well studied in recent years.
%, since it is an essential step of many subsequent face-related applications, including face alignment, face verification and recognition. 
From the pioneering work of Viola-Jones face detector~\cite{viola2001rapid} to recent state-of-the-art CNN-based methods, the performance of face detectors has been improved remarkably. For example, the average precision has been boosted to over 98$\%$~\cite{hu2016tinyface,najibi2017ssh,zhang2017s3fd} in the unconstrained FDDB dataset.

Although face detection algorithms have obtained quite good results under general scenarios, detecting faces in specific scenarios is still worth studying. 
For instance, one of the remaining challenges is partially occluded face detection. Facial occlusions occur frequently, e.g. facial accessories including sunglasses, masks and scarfs. %The detection of occluded faces is indispensable in some applications, such as video 
\begin{figure}[t]
	\begin{center}
		\includegraphics[width=\linewidth]{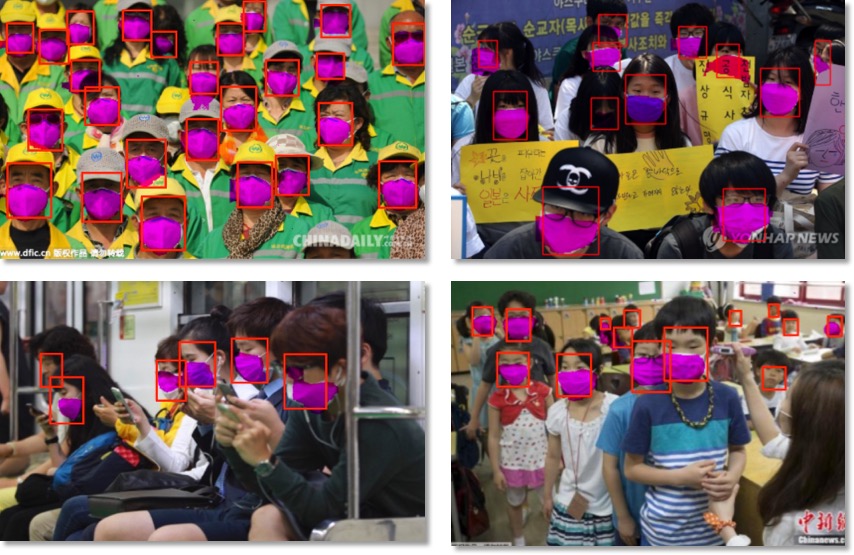}
	\end{center}
	\vspace{-6pt}
	\caption{The proposed AOFD is able to detect various heavily-occluded faces, while the occlusion-aware segmentation branch predicts masks of the occluded area. }
	\vspace{-10pt}
	\label{fig:cover}
\end{figure}
%surveillance and criminal investigation.
Occluded faces are only partially visible, and occluded regions have arbitrary appearances that may diverse from normal face regions. Hence occluded faces have significant intra-class variation, leading to difficulties in learning discriminative features for detection.
A standard paradigm to address this problem is to enlarge the training dataset of occluded faces, but it can't solve this problem in essence. Moreover, the lack of large-scale occluded face datasets makes it harder to handle this obstacle.

In this paper, we propose a framework for occluded face detection, aiming at formulating a new strategy to tackle the problem of limited occluded face training data, and exploiting the power of CNN representations for the faces with occlusions as far as possible. Firstly, motivated by the remarkable success achieved by adversarial learning in recent years, a deep adversarial network is proposed in our approach to generate face samples with occlusions from the mask generator. A compact constraint is adopted to reinforce the realness of generated masks. Secondly, we introduce an occlusion-aware model by predicting the occlusion segments at the same time with detecting faces. %Since occlusions not only impact the face detection process, but also severely deteriorate the performance of subsequent face alignment and recognition processes. It will be very meaningful to determine whether occlusion exists in a detected face for subsequent face-related applications. The predicted occlusion segment can be an important supervisory signal in occluded face alignment as well as occluded face recognition. 
%For detection task, a segmentation branch can be of great help in locating heavily-occluded face area.

In all, the generator aims to make the model focus more on the exposed areas, while the segmentation branch is to extract more informative features of the occluded area.
Intuitively, jointly solving these two tasks can be reciprocal.

To sum up, we make contributions in the following aspects:
\begin{itemize}
  \item A novel adversarial framework is proposed to alleviate the lack of occluded training face images by generating occluded or masked face features. We employ a compact constraint to get more realistic occlusions.
  \item Mask prediction is conducted simultaneously while detecting occluded faces. The occluded area will NOT be regarded as a hindrance but an auxiliary of face detection.
  %which is pretty meaningful for not only face detection but also other subsequent face-related tasks. 
  %Comprehensive analysis shows that mask prediction is helpful for developing robust occluded face detectors.
  \item 
  Experimental evaluations on the MAFA dataset demonstrate that the proposed AOFD can significantly improve the face detection accuracy under heavily occlusions. Besides, AOFD can also achieve competitive performance on the unconstrained face detection benchmark.
%  The proposed AOFD achieves competitive performance on the unconstrained face detection benchmark. Besides, experimental evaluations on the MAFA dataset also demonstrate the superiority of our method for partially occluded face detection.
\end{itemize}

\begin{figure*}[t]
	\begin{center}
		\includegraphics[width=\linewidth]{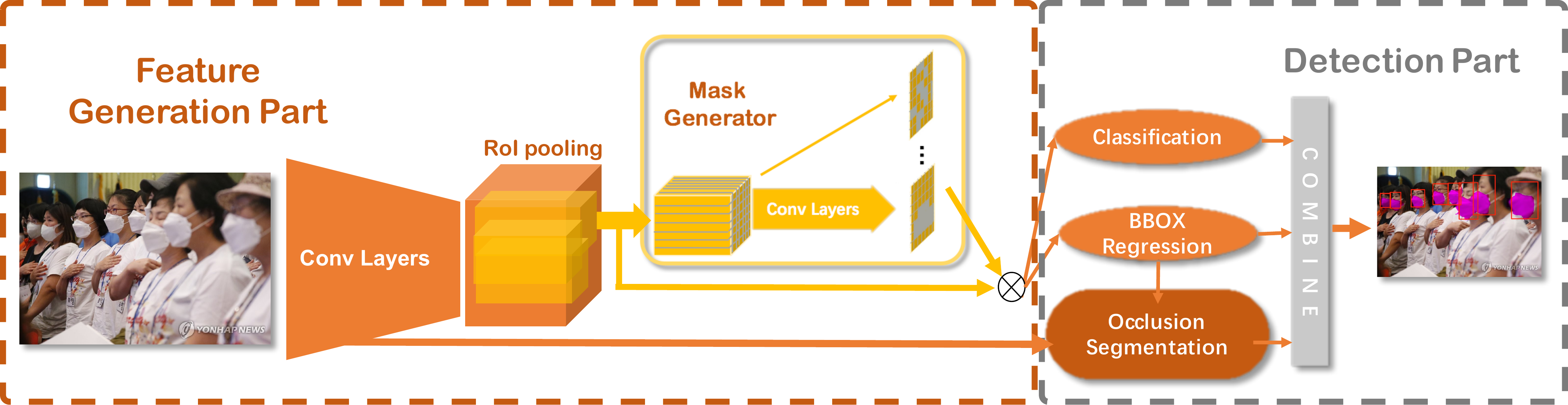}
	\end{center}
	\vspace{-6pt}
	\caption{The overall architecture of the proposed method. Mask generator are operated directly on RoIs where one type of masks are selected for each proposal.}
	\vspace{-10pt}
	\label{fig:model}
\end{figure*}

\section{Related Work}

We first briefly survey face detection algorithms, followed by a review of the state-of-the-art occluded face detection researches.
%have attempted to tackle face detection by exploring successful algorithms for general object detection.

Face detection can be considered as a special task of object detection. Successful general face detection algorithms often show great performance on face recognition. The Viola-Jones~\cite{viola2001rapid} detector can be recognized as a milestone in the field of face detection. They innovatively adopted AdaBoost to train cascade classifier with Haar-like features, which first makes it possible to apply face detection in real-time applications. Following their work, lots of boosting-based models were proposed~\cite{li2013learning,dollar2014fast,mathias2014face,zhu2016group}, focusing on designing more sophisticated hand-crafted features or improving the boosting strategy. More Recently, CNN features~\cite{yang2015convolutional} were utilized in this boosting framework.
%Another famous category of face detectors is DPM-based
Another famous category of face detectors is DPM-based. Deformable part models~\cite{felzenszwalb2010DPM} were proposed for object detection at first, which acquired impressive accuracy in complex environment. Inspired by this model, many extensions of DPM were developed to face detection~\cite{ghiasi2014occlusion} by modeling potential deformations among facial parts.
However, DPM models suffered from the high computational complexity, making it difficult to be applied in real-world applications such as digital cameras, phones or other mobile devices.

Generally speaking, boosting-based methods and DPM-based methods design features and optimize classifiers separately. The pipeline of these methods is divided into two stages, which is not an end-to-end architecture. Recently, benefitting from the prosperity of social network and big data, numerous deep learning based object detection algorithms have been proposed~\cite{girshick2015fast,ren2015faster,10.1007/978-3-540-74549-5_10,liu:inria-00325656,DBLP:journals/corr/ChenL17d}. CNN-based detectors therefore have become the mainstream in face detection gradually~\cite{farfade2015multi,li2015convolutional}. CNN-based face detectors directly learn robust face representations from data and optimize classifiers in an end-to-end style. For example, \cite{zhang2016joint} developed a deep cascaded multi-task framework that predict face and landmark location in a coarse-to-fine manner, and \cite{yu2016submodular} further improved the performance of cascade models by optimizing feature selection algorithms.

Although many efforts have been made in face detection, the performance of occluded face detection is still far from satisfactory, and there are few works on occluded face detection as far as we know. \cite{yang2015facial} explicitly inferred faceness score through local part responses via an attribute-aware model. But additional face-specific attribute annotations needed in this method were very difficult to collect. 
\cite{opitz2016grid} introduced a specific grid loss layer into CNNs that minimized the error rates on each sub-block of the feature map independently, thus every sub-part is discriminative on its own. 
%proposed a novel grid loss layer for CNNs to deal with partial occlusion in face detection, which minimized the error rate on sub-blocks of a convolution layer independently rather than over the whole feature map.
\cite{mahbub2016partial} introduced a partial face detection approach based on detection of facial segments. They mainly focused on detecting incomplete faces that captured by the front camera of smart phones. Recently, \cite{ge2017detecting} combined pre-trained CNN features with local linear embedding (LLE-CNN) to get similarity-based descriptors for partially visible faces. They built a dataset for masked face detection specifically, named the MAFA that contains 35K occluded faces. %.is the first dataset that was developed especially for masked face detection.
\cite{wang2017fan} applied anchor-level attention on Feature Pyramid Networks \cite{DBLP:journals/corr/LinDGHHB16}.

%The challenge of masked face detection can be ascribed to two aspects: 1) lacking of large-scale occluded face training data, and 2)
As mentioned above, our work is also related to adversarial learning. Generative Adversarial Network (GAN)~\cite{goodfellow2014generative} has shown great performance in numerous computer vision applications including image style transfer~\cite{zhu2017unpaired,pix2pix2016}, image generation~\cite{shrivastava2016learning,huang2017beyond} and so on. Adversarial learning provides a simple yet efficient way to train powerful models via the min-max two-player game between the generator and the discriminator. Most of the previous work focused on promoting generators. Recently, researchers began to pay attention to increase the capacity of discriminator by adversarial learning. \cite{wang2017fast} used adversarial learning in generating hard examples for object detection. \cite{li2017perceptual} employed Perceptual GAN to enhance the representations for small objects. Inspired by these applications, we develop an adversarial occlusion-aware model, which can synthesize occlusion-like face features for boosting occluded face detectors.

\section{Methods}
%The proposed model AOFD provides an effective and laconic method to tackle one of the most common and vicious problems in face detection-occlusion. This section first analyzes the problem (Sec. 3.1) and summarizes the overall architecture (Sec. 3.2), and then introduces our mask generation method (Sec. 3.3) and segmentation method (Sec. 3.4) separately.

In this section, we propose an AOFD method to tackle one of the most common and vicious problems in face detection-occlusion problem. We first analyze the occluded face detection problem (Sec. 3.1) and summarizes the overall architecture of AOFD (Sec. 3.2), and then introduce the mask generation and segmentation method in AOFD in Sec. 3.3 and Sec. 3.4, respectively.

\subsection{Problem Analysis}\label{sec:problemanalysis}
In real-world situations, we can generally classify face occlusion problems into three categories: facial landmark occlusion, occluded by faces and occluded by objects. 
%As demonstrated in Figure \ref{fig:occlusions}, 
Facial landmark occlusion includes conditions like wearing glasses and gauze masks. Occluded by faces is a complicated situation because a detector easily mis-recognize several faces into one or only detect a part of the faces. The segmentation method is proposed in order to mitigate this problem. When occluded by an object, usually more than half of a face will be directly masked. An original masking strategy is used to mimic these in-the-wild situations.

\begin{figure*}[t]
\centering     
\subfigure[]{\label{fig:a}\includegraphics[width=0.63\linewidth]{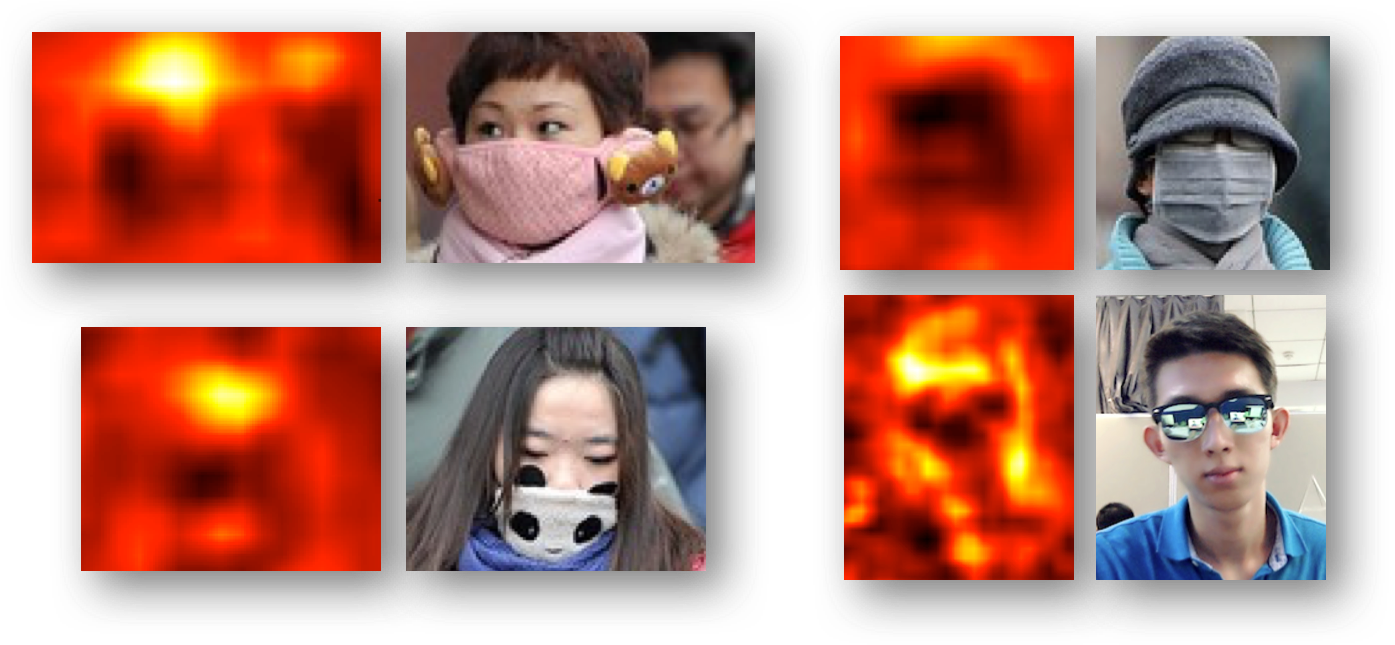}}
\subfigure[]{\label{fig:b}\includegraphics[width=0.3\linewidth]{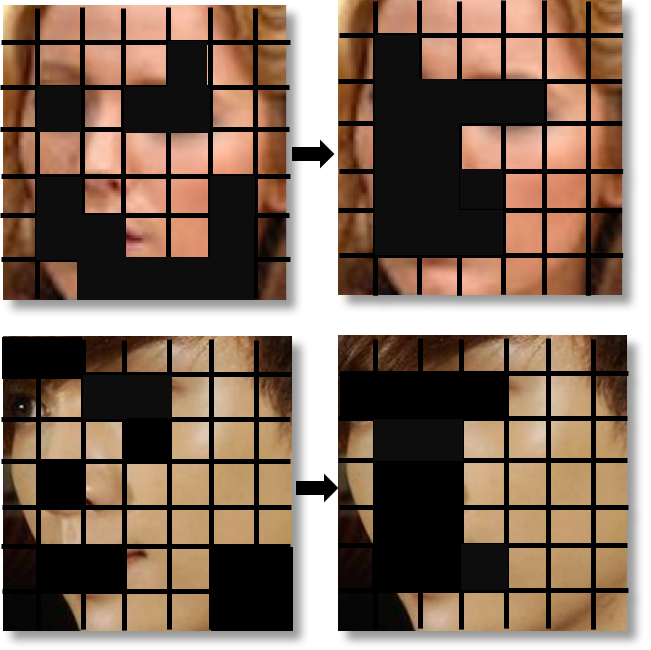}}
\caption{\textbf{(a)} Heat map of the features from conv5\_3 extracted from Faster RCNN. Occluded areas have less information in the features (the black areas). \textbf{(b)} The compact constraint has made the generated masks more accurate and efficient. A quarter of the minimum values is selected as the mask and is marked black. The masks are operated directly on the 7 by 7 RoIs. We map these masks to their corresponding receptive fields in the original image space for a better visualization. }
\label{fig:twopic}
\end{figure*}

We also visualized features of occluded faces in Figure \ref{fig:twopic} (a), finding that occluded areas rarely respond. For some heavily occluded faces, useful information in feature maps is too scarce for a detector to identify. To tackle this problem, we may need to enhance representation ability of exposed area. Meanwhile, recognition of occluded area can also bespeak that ``there is a face'' on the condition that sufficient context information is provided. For the most complex problem where a face is occluded by another face, the context area should cover at least the nearby faces and a larger receptive field is required so that the integrity of the background information can be ensured. This idea is enlightened by human vision, that is, human need a large context to define a small or incomplete object. Besides, as the quality of features directly determines the results, segmentation is better conducted on image features than on RoIs in order to extract more informative feature maps.

\subsection{Overall Architecture}
In order to detect faces with heavy occlusion, Adversarial Occlusion-aware Face Detector (AOFD) is designed with the view of (1) effectively utilizing the exposed facial areas, and (2) transferring the interference of the occlusions into beneficial information. For the first problem, we find that undetected faces are typically those with their characteristic part of face occluded, such as eyes and mouth. 
%In some cases, for example, sun glasses and gauze mask cover even all the face area, making it harder to detect. 
One feasible way is to mask the distinctive part of face in training set, forcing the detector to learn what possibly a face looks like even if there is less exposed area. To this end, a mask generator is designed in an adversarial way to generate a mask for each positive sample. It will generate different masks with faces of different poses. A masking strategy is applied for a better utilization of the mask generator as well. More details are illustrated in Sec. 3.3.

For the latter problem, We believe that finding common occlusions is helpful to detect incomplete faces behind them. Thus, an occlusion segmentation branch is introduced to segment occluded areas including hair, glasses, scarves, hands and other objects. This is not an easy task due to few training samples. Therefore, we labeled 374 training samples downloaded from internet for occlusion segmentation and came up with an original training strategy. This dataset is denoted as SFS (small dataset for segmentation). More details are listed in Sec. 4. 
%The name SFS (small dataset for segmentation) will be used in this paper to denote our training images for segmentation. %One thing to be noted is that we are not aiming to get a segmentation result with extremely high accuracy, the meaning of this branch is to let the detector know what possibly an occlusion in front of a face looks like, and thus facilitates heavily occluded face detection.

As is demonstrated in Figure \ref{fig:model}, a mask generator is added after a region of interest (RoI) pooling layer, followed by a classification branch and a bounding box regression branch. Finally, a segmentation branch is in responsible to segment the occluded area inside each bounding box. The final result combining classification, bounding box regression and occlusion segmentation will be output in the end. The overall loss of our architecture takes the following multi-task form:

% \textbf{Loss function:} The overall loss is a multitask loss:
\begin{equation}
L = \alpha L_c + \beta L_b+ \mu L_s
\end{equation}
where $L_c$ denotes a binary softmax loss for classification, $L_b$ denotes a smooth L1 loss for bounding box regression. We apply a binary softmax loss for segmentation branch, which is $L_s$. During training, the coefficients $\alpha$, $\beta$ and $\mu$ are set 1, 1, and 1$e^{-5}$ respectively. 

\subsection{Mask Generator}
 \textbf{Mask generator:} Since human face is very structural, facial features tend to appear in similar locations. However, with different poses, expressions and occlusions, distinguished facial area varies significantly. Our aim is to find this distinguished area and to generate a customized mask. We visualize some mask examples in Figure \ref{fig:twopic} (b). As we have observed, occluded area in features rarely respond in real images. To simulate this characteristic, masks are directly operated on RoIs. Therefore, the generator, which contains four convolutional layers with a straight mapping, is designed simply as it can be regarded as a binary prediction problem. Besides, the peculiarity of our mask generator, to distinguish from ~\cite{wang2017fast}, is the original generating procedure and the mask forms. Since face structures are inherently different from those of objects, they need to be learned in a more subtle and flexible way, or no plausible mask can be obtained.

% \begin{figure}[t]
%	\begin{center}
%		\includegraphics[width=\linewidth]{figures/3.png}
%	\end{center}
%	\vspace{-6pt}
%	\caption{The compact constraint has made the generated masks more accurate and efficient. A quarter of the minimum values is selected as the mask and is marked black. The masks are operated directly on the 7 by 7 RoIs. We map these masks to their corresponding receptive fields in the original image space for a better visualization.}
%	\vspace{-10pt}
%	\label{fig:masks}
%\end{figure}

 \textbf{Masking strategy:} The generated mask is a one-channel heat map where 0 represents masked area and 1 otherwise. During training, each pixel value will be squeezed to zero or one. We select a quarter of the minimum values as the mask when training the generator and one-third of the minimum values when training the overall model. 

Heavily occluded samples after masking will become an extremely hard source for training, making the model difficult to converge. To this end, three types of masks are proposed and jointly training with the original features. The first type is to use mask generator, which corresponds to facial landmark occlusion. The second type is to mask half of the features, whether left, right, top or bottom, and the third type is randomly dropping half of the pixels. This masking strategy embodies in-the-wild occlusion types analyzed in Sec. 3.1.

 \textbf{Loss function: }When training the mask generator, we employ an adversarial training method. We aim to increase classification loss as much as possible. Since a masked area is limited and a distinguished facial area is comparatively salient in feature maps, the model can easily converge. However, we find it not enough because the occluded area is sometimes strip-like or sporadic, while it is supposed to be more compact in real situations. Recall that the areas with longer or irregular edges will have a larger value for each pixel using a kernel of an edge detector. A kernel to make the occluded area sleeker and more circular is designed as a compact constraint for generated masks. The loss function is:
\begin{equation}
L_g = \gamma L_{com} - \eta L_c
\end{equation}
where $L_g$ denotes the loss for generator, $L_{com}$ denotes a compact loss, and $\gamma$ and $\eta$ are coefficients. $\gamma$ is set 1$e^{-6}$ and $\eta$ is set 1 in order to balance the derivatives. The compact loss is computed with a convolutional layer in a way as follows:
\begin{equation}
L_{com} = \sum((1-mask) \ast kernel)
\end{equation}
where $\ast$ denotes a convolutional operation, $mask$ is the first type of mask generated by the mask generator and the last item is the designed kernel, which is
$$
\left[
\begin{array}{ccc}
 	-\frac{1}{8} & -\frac{1}{8} & -\frac{1}{8} \\
	-\frac{1}{8} & 1 & -\frac{1}{8} \\
	-\frac{1}{8} & -\frac{1}{8} & -\frac{1}{8}
\end{array}
\right]
$$
In this way, strip-like or sporadic areas will get very high penalty and more reasonable masks can be obtained.

\subsection{Segmentation}
 \textbf{Design:} Previous works on segmentation have proved that CNNs is capable of comprehending the semantic information of a picture and elaborately conduct a pixel-wise classification. When combining detection with segmentation, it is usually designed in RoI level to achieve higher accuracy. 
 
Considering segmenting each RoI, one problem in occluded face situation is that the overlap of two bounding boxes will have different meanings. For example, if one face is occluded by another face, part of the front face should be regarded as an occlusion for the back face, while there shouldn't be any occluded area for the front face. Since our destination is to utilize the effective information contained in the occluded area to confirm if there is a back face and then make the exposed area more distinguished, ample context information is required. Moreover, it is the features that really matter in the bounding box classification and regression branch. With reasons above, segmentation is conducted in image level to directly affect the image feature maps. Therefore, the detector is able to find faces with more informative features embodying image-level signals like the appearance of an occlusion or a person. We call this method as an occlusion-aware method.

\begin{figure}[t]
	\begin{center}
		\includegraphics[width=\linewidth]{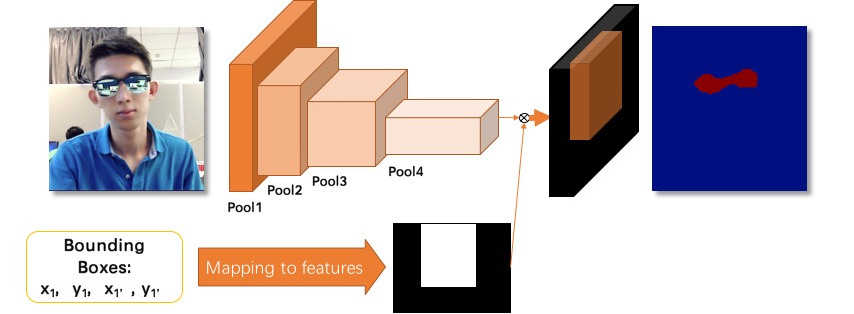}
	\end{center}
	\vspace{-6pt}
	\caption{The image segmentation branch in AOFD. Only areas inside bounding boxes are maintained so as to reduce noise.}
	\vspace{-10pt}
	\label{fig:segmentation}
\end{figure}

The segmentation branch is designed in a fully convolutional way (Figure \ref{fig:segmentation}). In order to obviate noise, it follows a bounding box regression branch and only areas inside bounding boxes are maintained. Bounding boxes are enlarged in scale with a factor of 1.3 before dropping the noise. Because it is only an auxiliary task for face detection, we didn't compare it with other segmentation methods. The results (Figure \ref{fig:cover}, Sec.4) should have shown the effectiveness. Although the final results have proved the feasibility of this method, the edges of segmentation seem to be a bit rough. This is caused by the limited size of the SFS training set. Nevertheless, we have verified the possibility to train the model with very limited training samples.

\textbf{Loss function:} We choose softmax loss instead of L1 or L2 loss used in some segmentation and image generation tasks~\cite{huang2017beyond,li2017perceptual} because it helps stabilize the training process.

%\begin{figure}[t]
%	\begin{center}
%		\includegraphics[width=\linewidth]{figures/demo.jpg}
%	\end{center}
%	\vspace{-6pt}
%	\caption{Qualitative results of AOFD on some hard samples from the internet. The proposed model can achieve impressive results on very hard examples.}
%	\vspace{-10pt}
%	\label{fig:demo}
%\end{figure}

\section{Experiments}
In this section, we qualitatively evaluate the proposed method with state-of-the-art methods. We first introduce detailed information during training (Sec. 4.1), and then test AOFD on several comparative benchmarks (Sec. 4.2). A series of ablative studies are conducted to verify the effectiveness of our method (Sec. 4.3). 
%Unfortunately, no dataset prepared for occluded face is available so far as the paper is written, and there exists a load of small faces in benchmarks, which is another urgent problem but not what our model is designed for. 
%As a matter of fact, Faster RCNN based methods generally have poor performance on small object detection. Simply applying more anchor scales and training image scales will doubtlessly increase the accuracy but also more computation and it is less insightful, so we are not doing this. However, the ability to detect large faces under all the conditions can also be considered a success. So the competitive results on FDDB~\cite{fddbTech} and MAFA~\cite{ge2017detecting} are capable enough to bespeak the superiority of AOFD. 
%Figure \ref{fig:demo} shows some qualitative results of our model.

\subsection{Training Details}
There are two stages in the training procedure. Firstly, based on Faster RCNN~\cite{ren2015faster}, we train the mask generator with the loss in equation (2). Secondly, the detector and the segmentation branch are trained jointly with the parameters in the mask generator fixed. 

In the second stage, due to the limitation of training data for segmentation, an unordinary training strategy is needed. We first train on SFS for 10k iterations, causing overfitting in segmentation. Then the model is trained on the combination of WIDER FACE training set and SFS for 50k iterations with loss weights for segmentation set 1$e^{-7}$ and finally the model is tuned only on SFS for 3 epochs. Derivatives from WIDER FACE training set will be zero for the segmentation branch during training. Because there are far more training images from the WIDER FACE training set than SFS, the segmentation branch can only be trained every several iterations while the features are changing all the time when training on the combined dataset. In this way, the overall loss can get rid of local minima and the overfitting problem can be solved. The basic learning rate is 0.001. AOFD runs 5 FPS on a TITAN X GPU, which is similar to the original Faster RCNN.

\textbf{Experiment settings:} AOFD is based on a Faster RCNN with a VGG16 backbone~\cite{simonyan2014very}. For anchors of PRN, we use three aspect ratios (1.7, 1 and 1.3) and four scales ($64^2$, $128^2$, $256^2$ and $512^2$). Batch size is set 1. An RoI is treated as foreground if its intersection over union (IoU) with any ground truth bounding box is higher than 0.5. To balance the number of foreground and background training samples, the ratio of foreground RoIs to background RoIs is set 1:3. During training, the short side of an input image is resized to either 512 or 1024 on condition that long side is no longer than 1024. 

\textbf{Small dataset for segmentation (SFS):} SFS contains 376 images with 1138 labeled faces downloading from the Internet. There is at least one occluded face in each image and over 80\% of the faces are occluded.

\begin{figure*}[t]
\centering     
\subfigure[Mask facilitates detecting]{\label{fig:a}\includegraphics[width=0.3\linewidth]{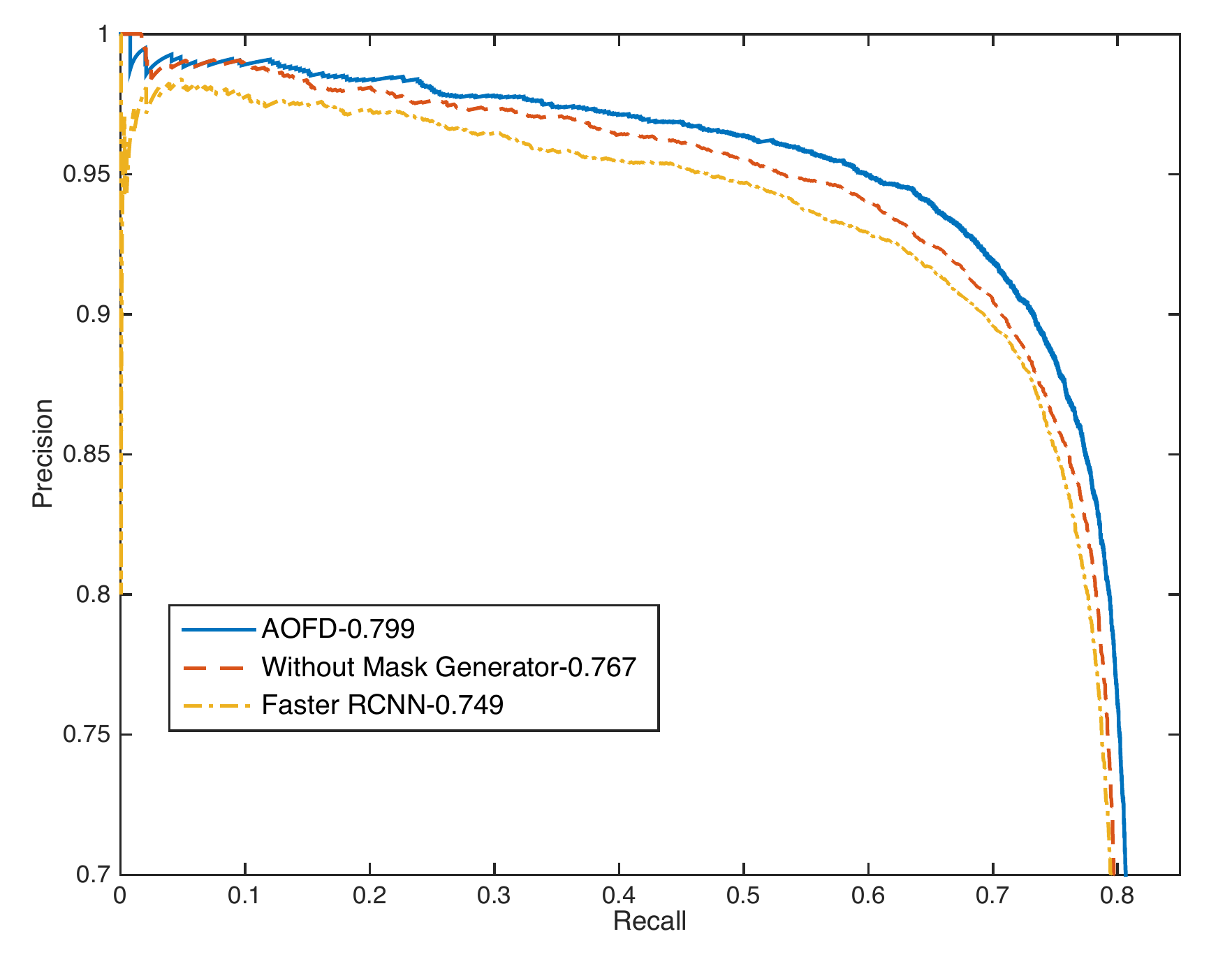}}
\subfigure[Segmentation increases recall]{\label{fig:b}\includegraphics[width=0.3\linewidth]{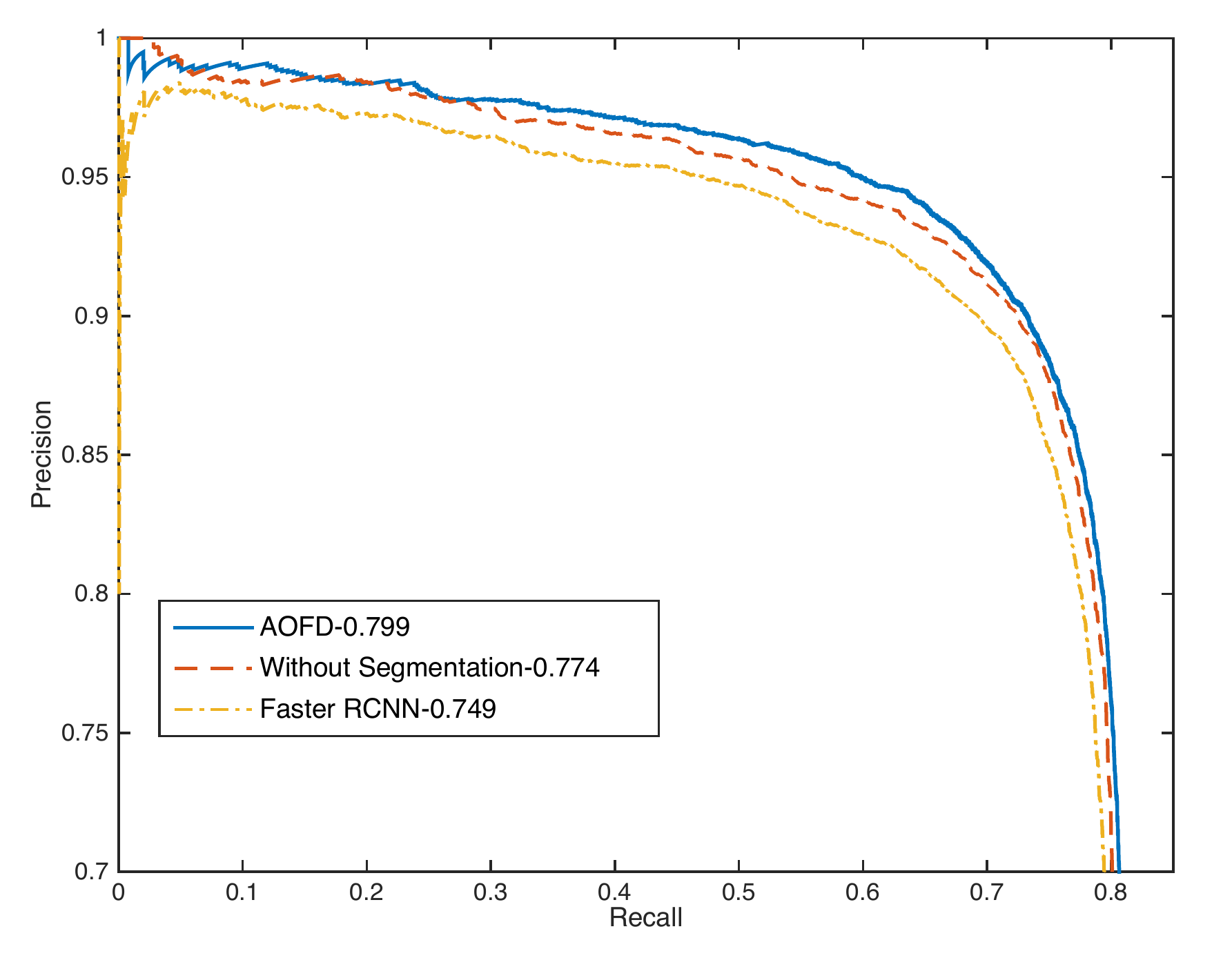}}
\subfigure[Mask area is crucial]{\label{fig:c}\includegraphics[width=0.3\linewidth]{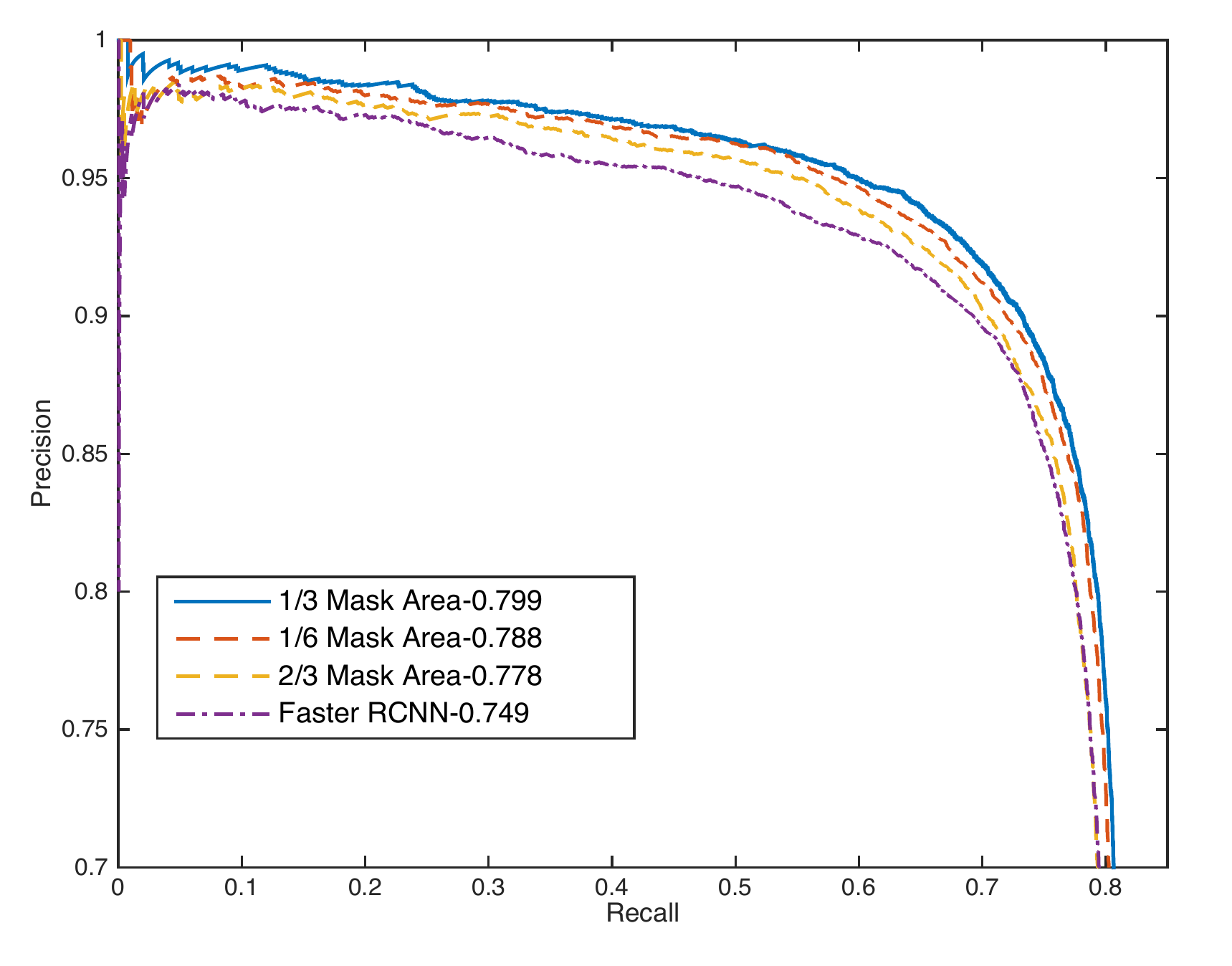}}
\caption{PR curves of ablative studies on the whole MAFA testing set without OHEM.}
\label{fig:ablativeROC}
\end{figure*}

\begin{figure*}[t]
\centering     
\subfigure[Discrete ROC curves]{\label{fig:a}\includegraphics[width=0.3\linewidth]{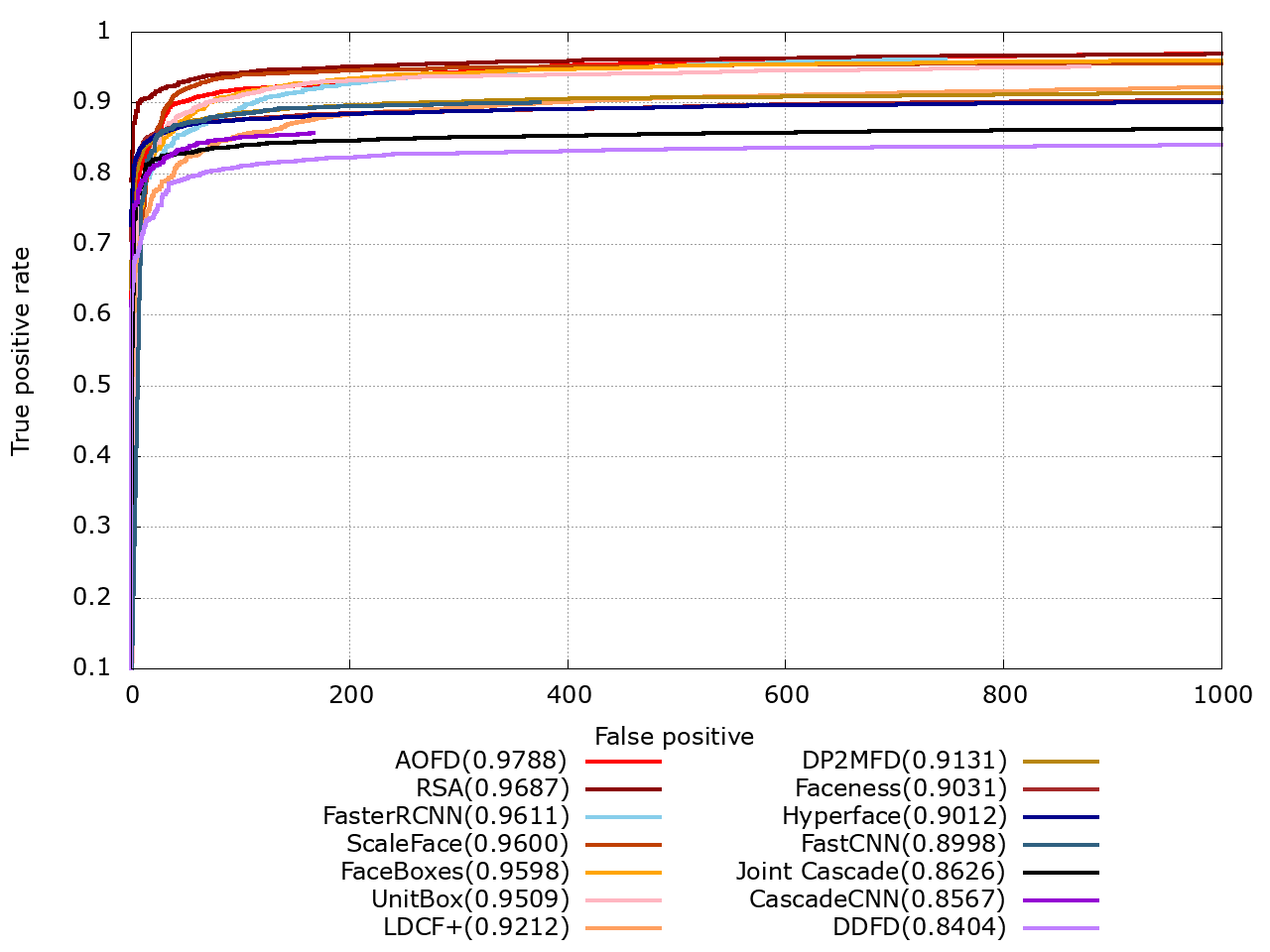}}
\subfigure[Continuous ROC curves]{\label{fig:b}\includegraphics[width=0.3\linewidth]{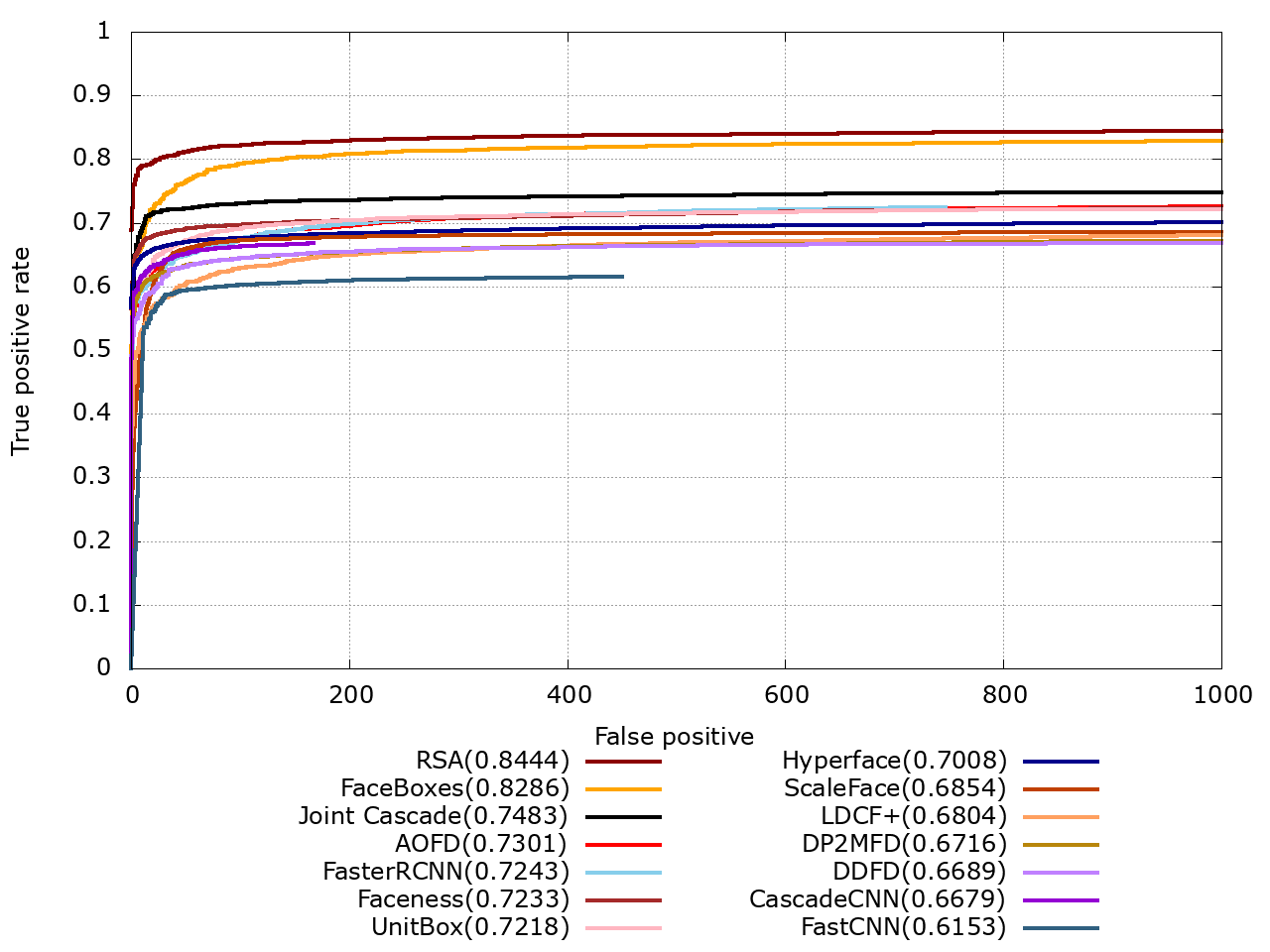}}
\subfigure[PR curves for our experiments on the MAFA testing set]{\label{fig:c}\includegraphics[width=0.3\linewidth]{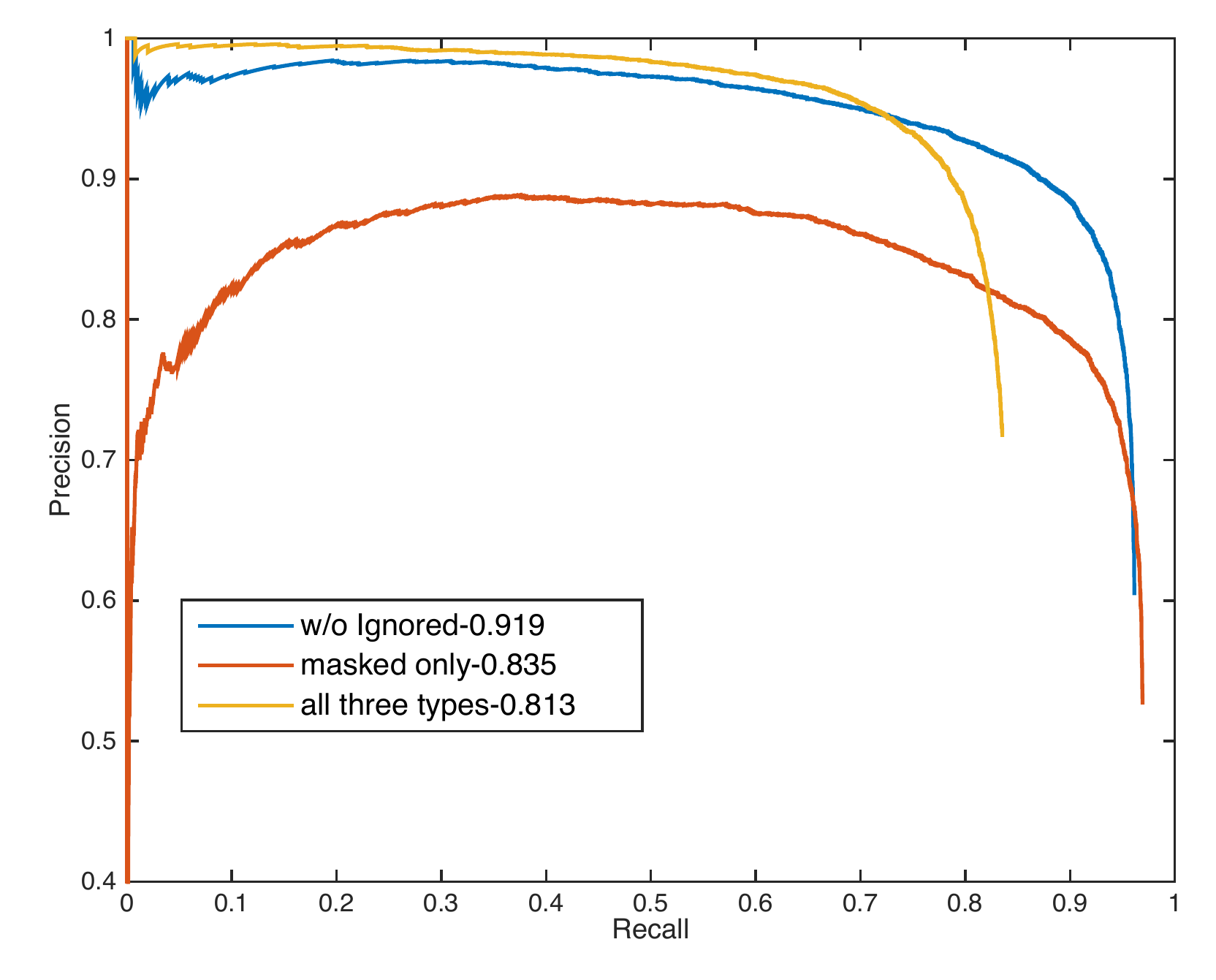}}
%\hspace{-6pt}
\caption{Results on FDDB ({\bf (a)(b)}) and MAFA testing set ({\bf (c)}).}
%\hspace{-10pt}
\label{fig:fddb}
\end{figure*}

\subsection{Evaluation on benchmarks}
Our model is trained on the WIDER FACE~\cite{yang2016wider} training set and evaluated on the FDDB and MAFA~\cite{ge2017detecting} databases. Although the MAFA database dose not release its training set, we still obtain state-of-the-art results on the MAFA testing set without fine tuning the model to adjust the variance between different annotation protocols. 
%WIDER FACE is not a desired source for testing since it contains half of tiny faces (the height range from 10 to 50 pixels) and our aim is not detecting tiny faces.
%\begin{figure}[t]
%	\begin{center}
%		\includegraphics[width=\linewidth]{figures/fddb3.jpg}
%	\end{center}
%	\vspace{-6pt}
%	\caption{Comparisons of several methods on FDDB test set. {\bf Top:} Discrete ROC curves. {\bf Bottom:} Continuous ROC curves.}
%	\vspace{-10pt}
%	\label{fig:fddb}
%\end{figure}

\begin{figure*}[t]
	\begin{center}
		\includegraphics[width=\linewidth]{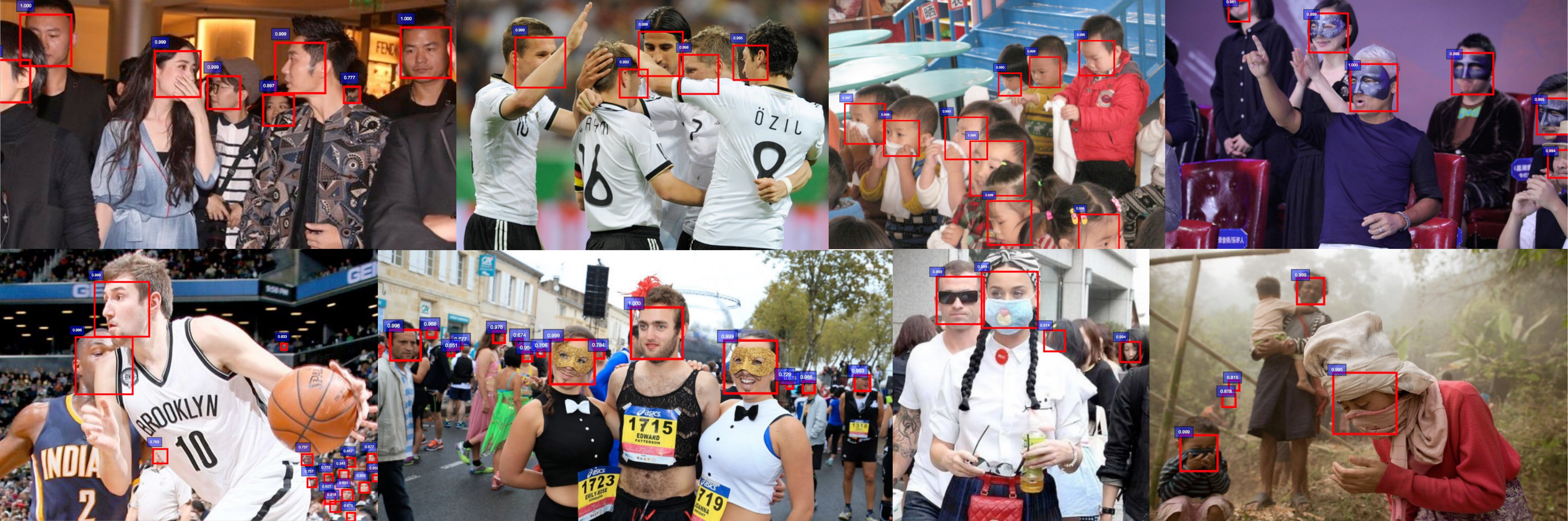}
	\end{center}
	\vspace{-6pt}
	\caption{Qualitative results of AOFD on the test set of the MAFA dataset.}
	\vspace{-10pt}
	\label{fig:mafademo}
\end{figure*}

\begin{figure}[t]
	\begin{center}
		\includegraphics[width=\linewidth]{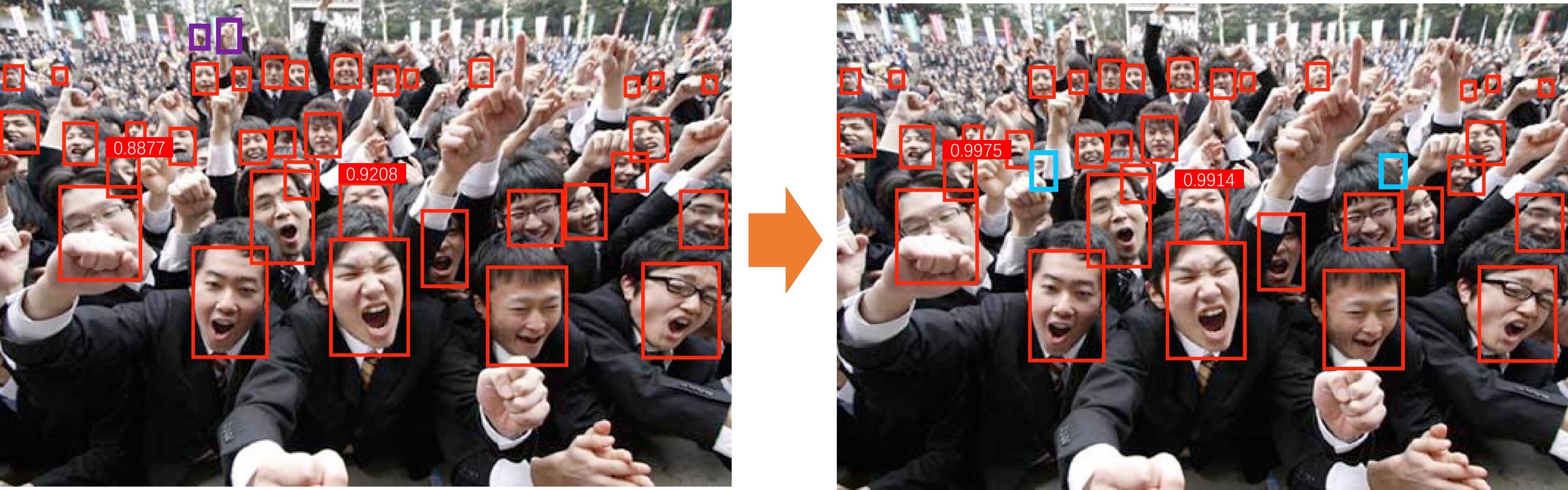}
	\end{center}
	\vspace{-6pt}
	\caption{Not only is AOFD able to detect occluded faces with higher confidences, but also capable of increasing the average precision. The {\bf purple} bounding boxes are false detections by  Faster RCNN and the {\bf blue} ones are new true positives detected by AOFD.}
	\vspace{-10pt}
	\label{fig:lift}
\end{figure}

\textbf{FDDB (Face Detection Data Set and Benchmark)} is an unconstrained dataset for face detection. It has 2, 845 images with 5, 171 faces. The detection results of different methods are shown in Figure \ref{fig:fddb}. 
%Noted that FDDB uses ellipse bounding boxes while ours are rectangle, so continuous score is negatively influenced but still fine. 
~\cite{liu2017recurrent} and several other methods obtain higher continuous score because they have transformed the rectangle bounding boxes into ellipse ones. 
%Besides, the criterion of the annotations from FDDB is slightly different from WIDER FACE which is our training source. For example, ellipse annotations in FDDB may include the whole head area while WIDER FACE annotations only refer to the face area. 
The fact that we didn't carry out extra training in the FDDB training set may lead to the increase of localization errors because of the difference of annotation criterion. However, in the comparison with state-of-the-art methods, we observe that AOFD outperforms all the other methods in terms of discrete score, demonstrating its strong ability to detect nearly all large faces even if faces with short side less than around 15 pixels are mostly neglected due to the anchor setting. 

Furthermore, our AOFD also obtains a higher recall rate at 1000 FPs on FDDB than other Faster RCNN methods with similar settings by a large margin (Figure \ref{fig:ablativeROC} and Figure \ref{fig:fddb}). 
%We can infer that it is true that CNN based models have capacity to learn some occlusions but far beyond enough and still quite a few occluded faces are omitted. Deliberately designed techniques for partially occluded face detection are required to further utilize the model's potential power.
The superior performance also reveals that applying masking strategy and training with a segmentation task are valuable attempts to enhance the model's capacity.

%\begin{figure}[t]
%	\begin{center}
%		\includegraphics[width=\linewidth]{figures/anno.pdf}
%	\end{center}
%	\vspace{-6pt}
%	\caption{Some examples of annotation in the MAFA testing set, where purple bounding boxes are labeled as `Ignored' due to blur, deformation or size, green bounding boxes are labeled as `masked' and blue ones are labeled as unmasked. We have covered all of these three types in the evaluation.}
%	\vspace{-10pt}
%	\label{fig:anno}
%\end{figure}

\textbf{MAFA} is designed for the evaluation of masked face detection, which contains 35806 face annotations with a minimum size of 32$\times$32. %The MAFA dataset covers 60 cases of masked faces in our daily scenarios concerning with 5 face orientations, 3 occlusion degrees and 4 mask types. 
Since the MAFA testing set uses squares to label faces, the rectangle bounding boxes in our results are transformed into squares to match the annotation.
%Experimental results demonstrate that AOFD achieves state-of-the-art performance on the MAFA dataset and outperforms other methods by a large margin.

There are three types of annotations in the MAFA dataset: masked, unmasked and ignored. Blurry or deformed faces or those with side length less than 32 pixels are labeled as `Ignored'. But we find that many `ignored' faces are also acceptable. 
%To fully demonstrate the capability of AOFD, we evaluated our model on the whole testing set. 
Since the other methods (~\cite{wang2017fan} ~\cite{ge2017detecting} ~\cite{zhang2016joint}) didn't count those annotations labeled as `Ignored', we report our results on both MAFA subsets and the whole testing set for comparison (Table \ref{tab:mafa}).

%\begin{figure}[t]
%	\begin{center}
%		\includegraphics[width=\linewidth]{figures/mafares.jpeg}
%	\end{center}
%	\vspace{-6pt}
%	\caption{PR curves for our experiments.}
%	\vspace{-10pt}
%	\label{fig:roc}
%\end{figure}

\begin{table}
	\renewcommand{\arraystretch}{1.5}
       \begin{center}
	\begin{tabular}{|l|c|c|c|}
		\hline
		Methods & All & `masked' only & w/o `Ignored' \\
		\hline\hline
		AOFD & \textbf{81.3\%} & \textbf{83.5\% } & \textbf{91.9\%} \\
		FAN & - & 76.5\% & 88.3\%\\
		LLE-CNNs & - & - & 76.4\%\\
		MTCNN & - & - & 60.8\%\\
		\hline
	\end{tabular}
	\end{center}
	\caption{Average precision on the MAFA testing set.}
	\label{tab:mafa}
\end{table}

As shown in Table \ref{tab:mafa}, the average precision achieves the highest 91.9\% (threshold 0.5) if we only evaluate on the faces with `masked' and `unmasked' labels. The result outperforms LLE-CNNs~\cite{ge2017detecting} by a large margin and is also better than the state-of-the-art Face Attention Network (FAN)~\cite{wang2017fan}. Since AOFD is proposed to address occlusion problem, we also evaluate our model on faces labeled as `masked' only. AOFD achieves 83.5\%  and has \textbf{7\% improvement} over the state-of-the-art result obtained by FAN.
% In some case studies, we find that the main type of false positives incurred by traditional methods are incomplete or unbalanced bounding boxes as shown in Figure \ref{fig:occlusions}. Our AOFD alleviates this problem to a great extend since it is aware of the occluded areas and can estimate preferable detections. 

Figure \ref{fig:fddb}(c) further shows the PR (Precision-Recall) curves of the three experimental settings. If we only count the faces annotated as `masked' (the orange curve in Figure \ref{fig:fddb}(c)), precision witnesses a sharp drop at the beginning. This is caused by unmasked and unlabeled face detections which are regarded as FPs when evaluating masked faces.
%since detector will find unmasked faces, which is regarded as FPs, with a higher priority when recall rate is low.
More results on MAFA are presented in Figure \ref{fig:mafademo}.

Furthermore, we have studied the main obstacle of our model to achieve a higher AP. The minimum IoU threshold for a true positive proposal is modified from 0.5 to 0.45, from which we observe that a slight decrease of IoU threshold can boost AP from 91.9\% to 93.8\%. This explains that the precision of bounding boxes can still be further improved

\subsection{Model Analysis}
To better understand the function of each part of our model, we ablate each component to observe AOFD's performance. In this way, the mask generator and segmentation branch are removed one after the another. We delve into the optimal area of the mask as well and find that the mask area is crucial for the functioning of mask generator. Besides, the efficiency of the compact constraint and the comparison with online hard example mining (OHEM)~\cite{shrivastava2016training} are also discussed in this section.

%Results on FDDB and MAFA test set are reported in Table ~\ref{tab:ablation} and Figure \ref{fig:ablativeROC}.

\begin{table}
	\renewcommand{\arraystretch}{1.5}
	 \begin{center}
	\begin{tabular}{|l|c|}
		\hline
		Settings & \multicolumn{1}{l|}{Recall rate at 1000 FPs} \\
		\hline\hline
		AOFD & 97.88\% \\
		%\hline
		w/o segmentation & 97.13\% \\
		%\hline\textsc{}
		w/o generator & 96.85\% \\
%		\hline
%		Mask Area (\(\frac{1}{6}\)) & 96.54\% \\
%		%\hline
%		Mask Area (\(\frac{1}{3}\)) & 97.12\%\\
%		%\hline
%		Mask Area (\(\frac{2}{3}\)) & 95.33\% \\
		\hline
	\end{tabular}
	 \end{center}
        \caption{Results of the ablative studies on FDDB.}
	\label{tab:ablation}
\end{table}

\textbf{Mask facilitates detecting:} State-of-the-art detectors are able to detect some of occluded faces, but with lower confidence. As shown in Figure \ref{fig:lift}, AOFD can increase the confidence of occluded faces by a large margin. Without the mask generator, AOFD pays less attention to exposed area or face structure, and the recall rate at 1000 false positives on FDDB drops by 1.3\%(Table \ref{tab:ablation}). The sharp decline (3.2\%) of average precision on the MAFA testing set in Figure \ref{fig:ablativeROC}(a) reveals the value of the mask generator as well.
It is also observed that AOFD's results would drop by around 1\% with only random and square-like occlusions. Since faces have unique structure characteristics such as facial symmetry, generating adaptive occlusions is essential in order to fool the detector.
%Their strategy is not suitable for face detection because masking only a square-like area is not enough to fool the detector in adversarial training, since faces have unique structure characteristics such as facial symmetry.

\textbf{Segmentation increases recall:} With the segmentation branch, the result in Table \ref{tab:ablation} witnesses an increase of 0.75\%. This improvement is relatively slight because there are not many heavily occluded faces in the FDDB testing set.
%confirming not only the effectiveness of this multitask method, but also our training strategy with limited data. We can also observe that 
The drop of average precision from 79.9\% to 77.4\% (Figure \ref{fig:ablativeROC}(b)) will be more convincing to confirm the effectiveness of the segmentation branch. 
%We should credit the mask generator for this commendable result.

\textbf{Mask area is crucial:} We find that the mask would vitiate the detector if a mask area is too large. Nevertheless, it would be of no use if it is too small. Figure \ref{fig:ablativeROC} gives a brief overview of our experiments, from which we find occluding one-third of features is an ideal area for a mask. 
%Noting that experiments with different settings all obtain state-of-the-art results. It is because only one of the three types of masks is influenced in the training strategy, and the other two remain effective.

\textbf{Compact constraint matters:} We propose a compact constraint $L_c$ to help generate more practical masks.
As is mentioned in Sec. 3.3, the generated masks are discrete or sporadic and are not plausible, e.g., two pixels occlusion on the mouth, three pixels occlusion on the eyes and others on the corners (Figure \ref{fig:twopic} (b)). In our initial experiments, the average precision is 0.785 when masking 1/3 of RoIs without the compact constraint, which is similar to having 1/6 masking area in Figure \ref{fig:ablativeROC}(c). However, masks become harder and more reasonable with $L_c$, which can account for the increase of performance.

\begin{table}
	\renewcommand{\arraystretch}{1.5}
       \begin{center}
	\begin{tabular}{|l|c|c|}
		\hline
		Methods & \multicolumn{1}{l|}{AP on MAFA} & Recall on FDDB\\
		\hline\hline
		single OHEM & 75.9\% & 96.54\%\\
		single AOFD & 79.9\% & 97.12\%\\
		AOFD with OHEM & 81.3\% &97.88\%\\
		\hline
	\end{tabular}%
	\end{center}
	\caption{Comparing AOFD with OHEM on the whole MAFA and FDDB testing set.}
	\label{tab:ohem}
\end{table}

\textbf{Comparing with OHEM:} 
%Online hard example mining (OHEM)~\cite{shrivastava2016training} has proved its efficiency by reranking and training hard examples in recent years, which has some similarity to ours-generating hard examples. 
We compare online hard example mining~\cite{shrivastava2016training} with our methods in Table \ref{tab:ohem}. We can see that the performance of training a Faster RCNN with OHEM is generally worse than a single AOFD without OHEM. 
%However, OHEM can assist in improving the performance. The results in MAFA and FDDB both witness a slight increase with OHEM.
But the combination of these two methods leads to a better performance. Although a harder training procedure means a more robust detector under this condition, the measurement of hard level needs to be carefully handled. For example, the decrease incurred by too large masking area demonstrated in Figure \ref{fig:ablativeROC}(c).

\section{Conclusion}
This paper has proposed a face detection model named AOFD to address the long-standing issue of face occlusions. A novel masking strategy has been integrated into AOFD to increase training complexity, and can plastically mimic different situations of face occlusions. The multitask training method with a segmentation branch provides a feasible solution and verifies the possibility to train an auxiliary task with very limited training data. The superior performance on both general face detection and masked face detection benchmarks demonstrates the effectiveness of AOFD.

{\small
\bibliographystyle{ieee}
\bibliography{btas}
}

\end{document}